\documentclass{article}

\usepackage[T1]{fontenc}
\usepackage{amsmath}
\usepackage{booktabs}
\usepackage{graphicx}
\usepackage{microtype}
\usepackage{url}
\usepackage{hyperref}
\usepackage{amsfonts}
\usepackage{amssymb}
\usepackage{tikz}
\usetikzlibrary{arrows.meta, patterns}

\definecolor{perlayer}{HTML}{0072B2}
\definecolor{ced}{HTML}{CC79A7}
\definecolor{cedswa}{HTML}{009E73}
\definecolor{shared}{HTML}{D55E00}
\definecolor{cachecolor}{HTML}{E69F00}

\usepackage[accepted]{mlsys2025}
\makeatletter
\renewcommand{\mlsys@appearing}{}
\makeatother

\graphicspath{{figures/}}

\newcommand{\Tau}{\mathrm{T}}
\newcommand{\R}{\mathbb{R}}
\newcommand{\MD}{M_{\mathrm{D}}}
\newcommand{\MDD}{M_{\mathrm{DD}}}
\newcommand{\MCED}{M_{\mathrm{CED}}}
\newcommand{\MCEDw}{M_{\mathrm{CED}\text{-}\mathrm{SWA}}}
\newcommand{\MLP}{f_\mathrm{MLP}}
\newcommand{\SA}{f_\mathrm{SA}}
\newcommand{\CA}{f_\mathrm{CA}}
\newcommand{\Head}{f_\mathrm{H}}
\newcommand{\Norm}[1][]{f_{\nu_{#1}}}
\newcommand{\bigO}[1]{\mathcal{O}(#1)}
\newcommand{\tran}{^{\mathsf{T}}}

\mlsystitlerunning{Context Compaction for World Models}

\begin{document}

\twocolumn[
\mlsystitle{Context Compaction for World Models}

\begin{mlsysauthorlist}
\mlsysauthor{Hayder Tirmazi}{bu,allspice}
\end{mlsysauthorlist}

\mlsysaffiliation{bu}{Boston University, Boston, Massachusetts, USA}
\mlsysaffiliation{allspice}{AllSpice Inc., Boston, Massachusetts, USA}

\mlsyscorrespondingauthor{Hayder Tirmazi}{hayder@bu.edu}

\mlsyskeywords{World Models, Context Compaction, Agent Harnesses, KV Cache, Long-Context Inference, Context Retrieval}

\vskip 0.3in

\begin{abstract}
World models let observers explore and interact with generated worlds. Recent
work shows that when observers revisit a region,
current models often fail to regenerate it consistently. We view this
as a context retrieval problem. LLM agents manage context retrieval at two
layers of their stack. At the model layer, the KV cache design determines how much
context fits in memory. At the agent harness layer, harnesses such as Claude Code pick the context sent to the LLM. Prior work on world models has mostly focused only on the model layer. In this paper, we make three contributions. First, we build a benchmark to measure world model context retrieval. It generates new worlds for every episode, preventing a
model from recalling rooms using training data. It also separates context retrieval from
guesses based on nearby frames. Second, we build an agent harness that performs context
compaction for world models, which to our knowledge no prior work has studied. Third, we
use our benchmark to study four world model designs.
Context compaction lets our world models retrieve rooms up to 4 times beyond their context
window without retraining. However, world models with
sliding-window attention lose most of their context retrieval when dreaming. Therefore, a dreaming world agent benefits more from extending the reach of its
world model using context compaction than switching to sliding window attention at the model layer.
\end{abstract}

]
\printAffiliationsAndNotice{}

\begin{figure*}[t]
  \centering
  \includegraphics[width=\textwidth]{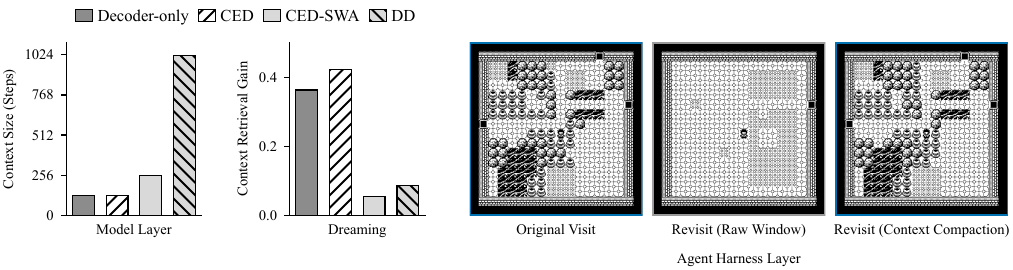}
  \caption{Left: Impact of model architecture on context
  retrieval for our video game room generation world model. Decoder-only needs 28.3 MB of KV cache memory for a context
  window of 128 steps. In the same memory, DD fits a context of 1,020 steps.
  Middle: Context retrieval gain of each world model when it dreams (with context compaction enabled). DD, which saves its memory
  with SWA, loses most of its context retrieval. Right: Agent harness layer can improve context retrieval on top of model layer changes. Decoder-only retrieves a room 
  revisited after 262 steps. The visit from 262 steps ago is no longer in Decoder-only's 128-step context window. The model therefore cannot reproduce the room. However, context compaction allows the original visit to still be sent to the model.}
  
  \label{fig:teaser}
\end{figure*}

\section{Introduction}\label{sec:introduction}

World models are systems that can predict a given environment from actions and observations. World models are useful in the realm of 
robotics and physical AI because they allow am RL agent to learn from a world model's imagined environment and then 
use that training to act in the environment~\citep{ha2018worldmodels,hafner2020dreamer}.
In addition to agents, people can also explore and play inside the vast and imaginative worlds constructed by world models~\citep{bruce2024genie} making way for a plethora 
of opportunities in the entertainment and video game industry~\citep{team2026alayaworld}.

When an observer explores and interacts with a world model's generated world 
for an extended period of time, they often return to places they visited earlier 
on in their trajectory. A realistic world model is expected to generate the same output for a given place on a revisit 
as it did on the first visit. Unfortunately, two recent evaluations~\citep{lian2025loopnav,ye2026mind} demonstrate
that world models do not currently maintain good consistency when observers revisit
the same area in the generated world.

We view this as a context retrieval problem, 
similar to how LLMs are expected to maintain the context of a user's conversation 
during long sessions. For LLMs, this is a well-explored area of research. 
Context in LLMs is typically managed at two layers of the inference stack: 
the model layer and the agent harness layer. At the model layer, 
there is a subtle tradeoff between the size of the KV cache and 
an LLM's context retrieval. If an LLM designer reduces the number of bytes 
of KV cache required per LLM token, it enables a fixed amount of GPU memory 
to fit a longer context. On the other hand, reducing the size of the KV cache
per token can also reduce how \textit{accurately} an LLM retrieves previous tokens in its context.

There are many optimizations that reduce the size of an LLM's KV cache. Multi-query attention~\citep{shazeer2019mqa} and Grouped-query attention~\citep{ainslie2023gqa} share an LLM's keys and values across 
multiple attention heads. A Decoder-Decoder model (DD)~\citep{sun2024yoco} computes an LLM's keys and values once for a token and shares them across all of the LLM's upper layers. Sliding Window Attention (SWA)~\citep{beltagy2020longformer} 
lets each of an LLM's tokens only attend to a fixed window of tokens preceding it. In addition
to the KV cache, some model architectures also focus 
on reducing the compute necessary for the prefill stage in inference. For example, 
recently, the Causal Encoder-Decoder (CED) introduced by DeepSeek-V4.1-Flash~\citep{deepseek2026ced}
lets an LLM project the keys and values for each upper layer from the hidden 
state of the middle layer. Note that the key and value weight matrices still remain separate 
for each upper layer in a CED, hence the KV cache in a CED is not smaller than the KV cache in a standard LLM. However, since the keys and values in the upper layers depend 
only on the middle layer, CED can fill the global KV caches of its upper layers during the prefill stage without running the upper layers on the full 
input prompt.

At the agent harness layer, a harness tries to manage the context of an LLM through
a process called context compaction~\citep{tirmazi2026compaction}. 
Most commonly, such as in Claude Code and OpenCode, context compaction involves 
creating an LLM-generated summary of the agent's session~\citep{tirmazi2026compaction}.
More recently, Codex with GPT-6 Astra can experimentally keep notes across context windows in
place of a single summary. Earlier context windows stay searchable, allowing the model
to retrieve details from previous messages and tool outputs when it needs them~\citep{openai2026astra}.

Unlike LLMs, the problem of context retrieval in world models has not been 
adequately studied. The vast majority of prior work is limited in scope
and focuses primarily on the model layer. Prior work
explores how a world model's architecture or its KV cache can be modified 
for better context retrieval. WorldMem~\citep{xiao2025worldmem} is a video 
world model that stores previously 
generated frames into a memory bank and retrieves them back into its diffusion 
transformer blocks during inference. Similarly, WorldPack~\citep{oshima2025worldpack} attempts to pack previously generated frames into 
fewer tokens allowing more frames to be fit into the context window during 
inference. WorldKV~\citep{yi2026worldkv} keeps 
and retrieve KV-cache chunks that are evicted by a sliding window. Finally, CaR~\citep{peng2026car} learns a compression network to improve context
retrieval.

All of the above papers evaluate their design by checking whether a frame that 
an observer revisits looks similar to the frame that the world model generated on the 
observer's original visit. However, this method of evaluating the 
model design does not
completely answer the isolated question of whether a world model is good at context 
\textit{retrieval}. When a world model generates a frame on an observer's revisit that 
is similar to the original frame, it can be 
due to three distinct reasons.
\begin{enumerate}
    \item The world model \textit{retrieved} the original frame from its \textit{context}.
    \item The world model \textit{reconstructed} a frame similar to the original frame 
    from its original \textit{training data} that the model memorized.
    \item The world model \textit{inferred} a frame similar to the original frame using 
    \textit{context clues} from \textit{other} recent frames that are in its context.
\end{enumerate}

In this paper, we call the latter two cases \textit{context reconstruction} and \textit{context inference} respectively, to distinguish them from \textit{context retrieval}. Since most evaluations of prior work score only the output of the world model, they cannot separate the three cases. The closest prior work to our approach is a recent study that replaces the memory that a world model reads with memory from elsewhere (such as another scene) and measures how much of the benefit of memory depends on its content \citep{tiwari2026audit}. However, they evaluate existing world models on existing datasets and do not control whether those scenes appear in the training data. Therefore, the study still cannot rule out context reconstruction.

When it comes to the agent harness layer, we are not aware of any prior work that uses an agent harness to manage the context of 
a world model similar to how existing LLM harnesses such as Claude Code and 
Codex manage the context of an LLM. The closest prior works we were able to find 
are WorldPlay~\citep{sun2025worldplay} and Echo-Memory~\citep{king2026echomemory}. 
While WorldPlay and the context-based variants of Echo-Memory use a retriever outside of the model to choose frames for a world model,
they require the world models to be trained using the frames chosen by the external retriever. 
In other words, they are not agent harness layer optimizations in the same sense 
as context compaction in LLM agents. Claude Code, for example, 
manages the context of an existing Anthropic LLM such as Claude Opus without having 
to change the LLM's weights or re-train it. World in World \citep{song2026worldinworld} retrieves earlier generated states for a frozen world model without training, but it does so inside the model, by restoring their KV caches and by rescaling attention during generation. Our agent harness leaves the computation of the world model unchanged and only decides which context the world model reads.

In this paper, we conduct a thorough study of the problem of context retrieval in world models. 
We train research world models of about 14M parameters that generate video game rooms and 
then use them to investigate how an array of systems decisions across the model layer and agent harness layer impact context retrieval. Concretely, we make the following three contributions.

\begin{enumerate}
    \item Create a benchmark for world models that specifically evaluates a 
    world model's context retrieval. Unlike prior work, our benchmark separates context retrieval from context reconstruction and context inference.
    \item Introduce an agent harness for world models that can perform context compaction. Our agent harness manages the context of an underlying world model without having to retrain or modify its weights. Our approach is similar to how agent harnesses for LLMs such as Claude Code, Codex, and OpenCode work.
    \item Conduct an evaluation of the effect of systems decisions at the model layer and agent harness layer (as well as their combinations) on context retrieval in world models, both when the world model reads true frames and when it dreams.
\end{enumerate}

We vary the model architecture and the agent harness independently 
and measure context retrieval on each configuration. Note that the two layers interact 
with each other and both play a part in a world model's ability to do accurate context 
retrieval. For example, a good context compaction algorithm at the agent harness layer
can extend the reach of a world model's context retrieval far beyond the model's context window.
Similarly, the design of the KV cache at the model layer significantly impacts the 
size of the model's context window assuming the available memory is fixed. Figure~\ref{fig:teaser} shows both effects. In practice, a world agent dreams, i.e., its world model predicts new frames from
frames that it generated earlier. Once our world agents dream, world models whose lower
blocks use SWA lose almost all of their context retrieval, with or without context
compaction, as we show in the middle plot of Figure~\ref{fig:teaser}.

\section{Background}
\label{sec:background}

Let $\MD$ be any decoder-only Transformer. We use $N$ to denote the number of transformer blocks in $\MD$. We use $D$ to denote the width of $\MD$'s embedding vectors, i.e., each token is mapped to a vector of length $D$. We use 
$\rho$ to denote the number of bytes used to store each floating point value in $\MD$. $\rho$ depends on the precision used to store $\MD$'s KV cache. Finally, we use $h$ and $d_h$ to denote the number of attention heads and the width of each attention head in $\MD$ respectively. Note that $h \cdot d_h = D$. Treating $D$ as a constant, the compute cost of
generating each new token after $L$ tokens in $\MD$ is $\bigO{N L^{2}}$ since
the compute cost of each individual transformer block in $\MD$ is $\bigO{L^2}$.

We can use a 
Key-Value (KV) cache to store the keys and values of each processed token 
in each transformer block in $\MD$. Using a KV cache reduces the compute cost
of generating each new token after $L$ tokens in $\MD$ to $\bigO{N L}$. $\MD$'s
KV cache needs to store $2 N L D \rho$ bytes. If we use Sliding Window Attention (SWA) with a window size of $w$ tokens then the compute cost of an SWA transformer block is $\bigO{w}$ and $\bigO{w^2}$ with and without a KV cache respectively. The KV cache only needs to store $2wD\rho$ bytes for an SWA transformer block. We derive these costs along with a more rigorous treatment of model architectures in Appendix~\ref{app:background}.

\paragraph{Decoder-Decoder Models.} We use $\MDD$ to denote a Decoder-Decoder Model~\citep{sun2024yoco}. In $\MDD$, $N$ transformer blocks are split into $N_s$ lower blocks and $N_c$ upper blocks, i.e., $N = N_s + N_c$. Following the SWA variant of YOCO, the lower blocks in 
$\MDD$ use SWA. Let $z$ be the normalized output of the last lower block in $\MDD$. $z$ is projected onto a shared key matrix $K^{\star} = z
W_K^{\star}$ and a shared value matrix $V^{\star} = z W_V^{\star}$. 
Each upper block in $\MDD$ computes its own queries and attends over $K^{\star}$ and $V^{\star}$ through cross-attention. This lets $\MDD$'s KV cache get away with only storing $2LD\rho + 2N_{s}wD\rho$ bytes. In other words, the part 
of the KV cache that grows with the model's history, $L$, is $N$ times smaller 
than in $\MD$ where it was $2NLD\rho$ bytes.

\paragraph{Causal Encoder-Decoder Models.} DeepSeek's recent 
model, DeepSeek-V4.1-Flash, uses a novel model architecture 
called a Causal Encoder-Decoder model~\citep{deepseek2026ced}. We use $\MCED$ to denote Causal Encoder-Decoder models. In an $\MCED$, similar to 
a Decoder-Decoder model, $N$ transformer blocks are 
split into $N_{s}$ lower blocks and $N_{c}$ upper blocks. The $N_{s}$ lower
blocks are called encoder blocks. They are ordinary decoder-only transformer blocks.
The $N_{c}$ upper blocks are called decoder blocks. 
Let $z$ be the normalized output of the last encoder block in $\MCED$. 
Each decoder block has a separate key weight matrix $W_K^{(i)}$ and 
separate value weight matrix $W_V^{(i)}$. The key $K^{(i)}$ and value 
$V^{(i)}$ for each decoder block are computed from $z$, i.e., $K^{(i)} = z W_{K}^{(i)}$ and $V^{(i)} = z W_{V}^{(i)}$. Note that the decoder blocks use 
cross-attention. As defined, $\MCED$ does not reduce KV cache size as compared 
to $\MD$, the decoder-only Transformer model. However, it does save compute cost.
Since each of the $N_{c}$ decoder blocks uses $z$, the compute cost of prefilling 
$L$ tokens in $\MCED$ is $\bigO{N_{s}L^2 + N_{c}L}$ compared 
to $\bigO{NL^{2}}$ in $\MD$. Our description of $\MCED$ follows the general formulation in the DeepSeek-V4.1-Flash technical report. However, note that 
DeepSeek-V4.1-Flash also adds some components orthogonal to the idea of Causal Encoder-Decoders which we do not 
include as part of $\MCED$ for simplicity.

\paragraph{Context Compaction.} There are two places where we can
store the history of the world being predicted by a world model. The first is the KV cache
within the model, which is common practice today. The second is the memory maintained by an external
agent or program that interacts with the model. While common in LLM agent harnesses, this is an underexplored concept in world models. An LLM agent harness typically
maintains a transcript of its state. When the transcript exceeds the context
window, the agent harness selects a subset or summary of the transcript to fit within
it. This is called context compaction~\citep{tirmazi2026compaction}.

\section{World Retrieval}

In this section, we introduce a benchmark that evaluates a world model's context 
retrieval capabilities in a way that decouples them from the world model's context reconstruction
and context inference capabilities, as discussed in \S~\ref{sec:introduction}.

\subsection{Benchmark World}\label{sec:benchmark-world}

\begin{figure}[t]
  \centering
  \includegraphics[width=\linewidth]{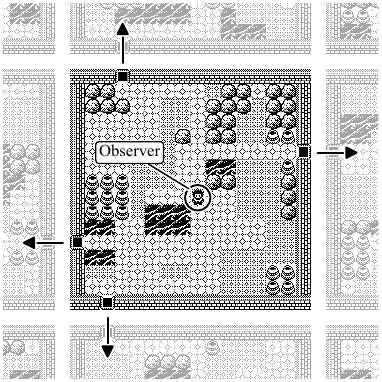}
  \caption{A frame in our benchmark world with the observer placed at the center of the room. The observer is allowed to use the doors to transition to the neighboring rooms in the world's $8 \times 8$ grid.}
  \label{fig:world}
\end{figure}

Our world consists of an $8 \times 8$ grid of video game rooms. Each room in the world is a $16 \times 16$ grid of tiles. There are eight types of tiles in the world such as a rock tile and a water tile.
The borders of the room hold walls 
and a door for each neighboring room. Consequently, each border can have at 
most one door tile and the rest of the tiles of the border are wall tiles. 
We use public domain sprites~\citep{drummyfish2018onebit} for the tiles. We uniformly randomly choose a room in the world and place the observer in the room. The observer 
is placed in the center of the current room and an action in the world 
consists of the observer moving through a door to a neighboring room. Consequently, each frame 
in our world is a given video game room. Similarly, each
step in our world is at the granularity of a room transition. Please refer to Figure~\ref{fig:world}.
In our benchmark, we generate episodes (for both training and evaluation) 
by making the observer navigate the benchmark world using a random walk. 
At any given step, we uniformly randomly choose one of the doors of the current room
and make the observer go through that door. However, we exclude the door 
the observer just entered through from the set of doors we uniformly randomly choose from. In case there is a dead end, i.e., when there is no door other than the door the observer entered from, we make the observer return through the same door.
Using our approach, the observer frequently revisits rooms without the need 
for scripted loops. The length of the walk between an observer's original visit and the observer's revisit to the same room ranges from a few steps to the length 
of the benchmark's episode. We denote this length as the \textit{gap} of the revisit for the remainder of this paper. We group room revisits by their gap into the following seven bins: 1--7, 8--15, 16--31, 32--63, 64--127, 128--255, and 256--511 steps. Note that we incrementally double the interval of the gap bins. 

\subsection{World Model}
\label{sec:designs}


Our benchmark world and its evaluation can be used with any world model that can 
ingest our video game rooms tokenized in any form and predict the next room. In this section, we provide an  overview of the world models we specifically train and evaluate on. For our world models, we use $8$ tokens to encode frames and $1$ token to encode 
the observer's action. We tokenize a video world room in the following way. Each of the $8$ tokens corresponds to a type of tile, e.g., the water tile is assigned a token. Each token is computed from a $16 \times 16$ boolean map where a $1$ at position $(x, y)$ corresponds to the water tile being at tile position $(x, y)$ in the video game room's $16 \times 16$ tile grid. The action token simply encodes the direction of the door the observer wants to transition to. Note that there is at most $1$ door for each of the four directions in a room. Consequently, our world models are sent $9$ tokens for each step. Our world models generally have a context window of $128$ steps. 

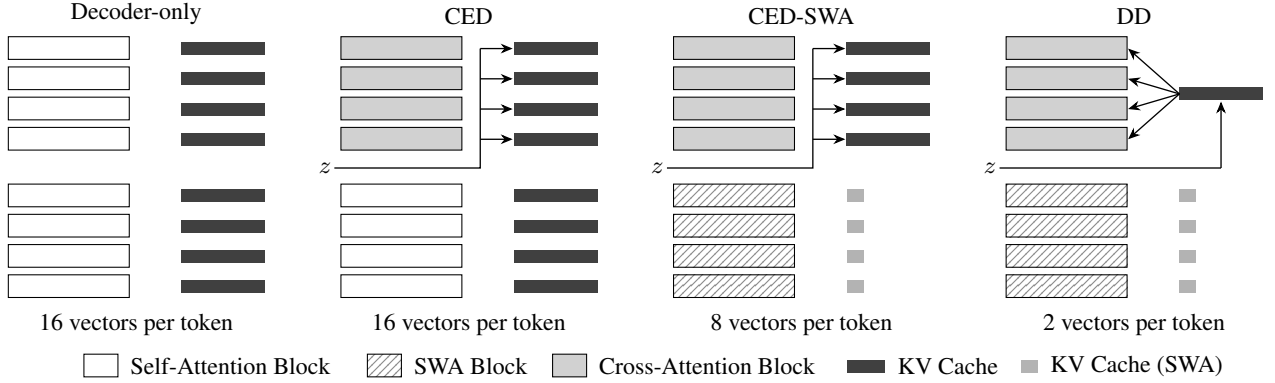
\begin{figure*}[t]
  \centering
\begin{tikzpicture}[
    x=1cm, y=1cm, font=\footnotesize,
    self/.style={draw, fill=white, minimum width=1.6cm, minimum height=0.30cm,
      inner sep=0, anchor=south west},
    swa/.style={self, pattern=north east lines, pattern color=black!45},
    cross/.style={self, fill=black!18},
    feed/.style={-{Stealth[length=1.6mm, width=1.3mm]}, line width=0.5pt},
    declare function={rowy(\r) = \r * 0.40 + (\r > 3) * 0.35;},
  ]
  \def\zy{1.72}
  \newcommand{\block}[3]{\node[#3] at (#1, {rowy(#2)}) {};}
  \newcommand{\fullcache}[2]{\fill[black!75] (#1 + 2.3, {rowy(#2) + 0.07}) rectangle ++(1.1, 0.16);}
  \newcommand{\windowcache}[2]{\fill[black!30] (#1 + 2.3, {rowy(#2) + 0.07}) rectangle ++(0.22, 0.16);}
  \newcommand{\projection}[1]{
    \node[anchor=east, inner sep=1pt] at (#1 - 0.08, \zy) {$z$};
    \draw[line width=0.5pt] (#1 - 0.08, \zy) -- (#1 + 1.85, \zy) -- (#1 + 1.85, {rowy(7) + 0.15});
    \foreach \row in {4, ..., 7} {
      \draw[feed] (#1 + 1.85, {rowy(\row) + 0.15}) -- (#1 + 2.28, {rowy(\row) + 0.15});
    }
  }
  \newcommand{\designlabel}[3]{
    \node[anchor=south] at (#1 + 1.7, {rowy(8) - 0.05}) {#2};
    \node[anchor=north] at (#1 + 1.7, -0.08) {#3};
  }

  \foreach \row in {0, ..., 7} {
    \block{0}{\row}{self}
    \fullcache{0}{\row}
  }
  \designlabel{0}{Decoder-only}{16 vectors per token}

  \projection{4.4}
  \foreach \row in {0, ..., 3} {
    \block{4.4}{\row}{self}
    \fullcache{4.4}{\row}
  }
  \foreach \row in {4, ..., 7} {
    \block{4.4}{\row}{cross}
    \fullcache{4.4}{\row}
  }
  \designlabel{4.4}{CED}{16 vectors per token}

  \projection{8.8}
  \foreach \row in {0, ..., 3} {
    \block{8.8}{\row}{swa}
    \windowcache{8.8}{\row}
  }
  \foreach \row in {4, ..., 7} {
    \block{8.8}{\row}{cross}
    \fullcache{8.8}{\row}
  }
  \designlabel{8.8}{CED-SWA}{8 vectors per token}

  \pgfmathsetmacro{\sharedy}{(rowy(4) + rowy(7)) / 2 + 0.07}
  \node[anchor=east, inner sep=1pt] at (13.2 - 0.08, \zy) {$z$};
  \draw[line width=0.5pt] (13.2 - 0.08, \zy) -- (13.2 + 2.85, \zy);
  \draw[feed] (13.2 + 2.85, \zy) -- (13.2 + 2.85, \sharedy - 0.02);
  \fill[black!75] (13.2 + 2.3, \sharedy) rectangle ++(1.1, 0.16);
  \foreach \row in {0, ..., 3} {
    \block{13.2}{\row}{swa}
    \windowcache{13.2}{\row}
  }
  \foreach \row in {4, ..., 7} {
    \block{13.2}{\row}{cross}
    \draw[feed] (13.2 + 2.3, \sharedy + 0.08) -- (13.2 + 1.62, {rowy(\row) + 0.15});
  }
  \designlabel{13.2}{DD}{2 vectors per token}

  \begin{scope}[shift={(1.0, -1.05)}]
    \node[self, minimum width=0.45cm] at (0, 0) {};
    \node[anchor=west] at (0.5, 0.15) {Self-Attention Block};
    \node[swa, minimum width=0.45cm] at (3.75, 0) {};
    \node[anchor=west] at (4.25, 0.15) {SWA Block};
    \node[cross, minimum width=0.45cm] at (6.2, 0) {};
    \node[anchor=west] at (6.7, 0.15) {Cross-Attention Block};
    \fill[black!75] (10.1, 0.07) rectangle ++(0.5, 0.16);
    \node[anchor=west] at (10.65, 0.15) {KV Cache};
    \fill[black!30] (12.4, 0.07) rectangle ++(0.22, 0.16);
    \node[anchor=west] at (12.67, 0.15) {KV Cache (SWA)};
  \end{scope}
\end{tikzpicture}
  \caption{The architecture of the four world models we train and evaluate 
  on. Each world model has $8$ transformer blocks. We show the KV cache 
  corresponding to a transformer block to the right of each respective block. From left to right, each architecture changes one property of the model incrementally. Switching from Decoder-only to CED, the upper blocks compute their keys and values from
  the output $z$ of the last lower block. From CED to CED-SWA, the lower blocks switch to using SWA so they only attend over the $2$ most recent steps. Finally, from CED-SWA to DD, the
  $4$ upper KV caches become one shared KV cache that is read by all the upper blocks.}
  \label{fig:architecture}
\end{figure*}

We train and evaluate on four world models. In the notation of \S~\ref{sec:background}, all four of our world models have $N = 8$ transformer blocks with $h = 6$ attention heads. Each attention head has width $d_h = 64$. 
All four world models also all have an embedding width of $D = 384$. We store the KV cache in float32, so we use $\rho = 4$ bytes per floating point value for all cache sizes. The first model uses the Decoder-only architecture, $\MD$, as described in \S~\ref{sec:background}. The second model uses the Decoder-Decoder architecture, $\MDD$. The third model uses a Causal Encoder-Decoder architecture, which we described using $\MCED$ with ordinary lower transformer blocks. Our fourth world model is a variant of $\MCED$ that uses SWA for its lower blocks similar to $\MDD$'s lower blocks. We will refer to this variant of $\MCED$ as $\MCEDw$.
Figure~\ref{fig:architecture} illustrates the design of the four 
world models. Note that all four models have roughly 14M parameters. 
Specifically, $\MD$
has 14,352,000 parameters. Both $\MCED$ and $\MCEDw$ have 14,352,768 parameters. Finally, $\MDD$ has 13,465,728 parameters. For simplicity, we use $\MD$, $\MCED$, $\MCEDw$, and $\MDD$ interchangeably with Decoder-only, CED, CED-SWA, and DD in the rest of this paper.

\subsection{Context Reconstruction}

Context reconstruction occurs if a world model has observed a frame in our world in its training data. For foundation world models that are typically pretrained on a large corpus of data from the internet and other sources, we prevent this by ensuring that the video game rooms in our benchmark world are randomly generated 
at benchmark time. For world models specifically trained for our benchmark world, 
we ensure that separate pseudorandom seeds are used for training and evaluation. Training on pseudorandom seeds of our benchmark world that are not the same as the evaluation seeds can certainly teach a world model general terrain statistics of 
the world such as the typical size of a patch of water. However, it cannot teach a world model the specific layout of a given room generated during evaluation. This property ensures that to predict a revisited room with a high accuracy upon a revisit, a world model cannot successfully rely on context reconstruction from its training data.

\subsection{Context Inference}\label{sec:context-inference}

Context inference occurs if a world model is able to accurately predict a revisited frame without actually retrieving the original visit frame from its context. As we discussed earlier, this can be due to the model using context clues from other recent frames that are in its context. To separate context retrieval from context inference, we score each revisit in two runs. In the first run, the 
world model reads its normal context window. In the second run, we keep only the $5$ most recent steps of the benchmark world in the context window and replace all earlier steps 
with the same number of steps from another benchmark world. We call the second run a \textit{shuffled} run and refer to its context window as a shuffled context window.

Note that the reason we chose to keep the $5$ most recent steps (as opposed to any other number of recent steps) is because $5$ steps is the receptive field of the SWA transformer blocks in two of the model architectures, CED-SWA and DD, that we describe in \S~\ref{sec:designs}. In these two model architectures, each of the four lower blocks attends over the $2$ most recent steps. The 
information from earlier steps therefore can move at most one step forward 
through each lower block. Hence the output of the final lower block depends on 
the five most recent steps only. If we keep the $5$ most recent steps, we can prevent
the input of the lower transformer blocks that rely on SWA from changing. The shuffled run therefore changes only the history that a model
reads through attention beyond its local window. We keep the same $5$ steps
for every model architecture in the paper, therefore, we score all model architectures on the same shuffled contexts.

Let $\alpha_{\mathrm{oc}}$ be the accuracy of a world model on a revisited room in our benchmark world when the world model is provided the original context window. Let $\alpha_{\mathrm{sc}}$ be the accuracy of the world model when it is provided a shuffled context window. We define the \textit{context retrieval gain}, $\Delta_{\mathrm{cr}}$, to be $\Delta_{\mathrm{cr}} = \alpha_{\mathrm{oc}} - \alpha_{\mathrm{sc}}$. 

For any room that was visited earlier than $5$ steps ago, if a world model is relying on context retrieval for its prediction, it will have a much lower $\alpha_{\mathrm{sc}}$ as compared to $\alpha_{\mathrm{oc}}$. This is because the shuffled context will not contain the originally visited frame that the world model is using to predict the revisited frame. Therefore, the world model's context retrieval gain will be large. On the other hand, if the world model is relying primarily on context inference, the difference between $\alpha_{\mathrm{oc}}$ and $\alpha_{\mathrm{sc}}$ will not be as stark, resulting in a lower context retrieval gain. 

We use revisits that span more than 128 steps after the original visit as a negative control. As we mentioned earlier, the world models we train and evaluate in this paper have a raw context window of 128 steps. A raw context window of 128 steps does not contain the original visit
at these gaps, even in the first run. Our negative control run (both with a regular context window and a shuffled context window) therefore does not have the original
visit in its context. Hence, the context retrieval gain for our negative control run 
should be close to zero. If we see a larger gain, it would imply that the shuffled 
run is failing to remove a helpful context clue. Our experiments show the negative control runs to have a context retrieval gain in the range $[0.03, 0.06]$ for all four world models serving as a validation for our approach of isolating context inference in our benchmark. Please see Appendix~\ref{app:gains} for more details.

\section{World Agents}
\label{sec:design}
\label{sec:harness}
\label{sec:compaction}

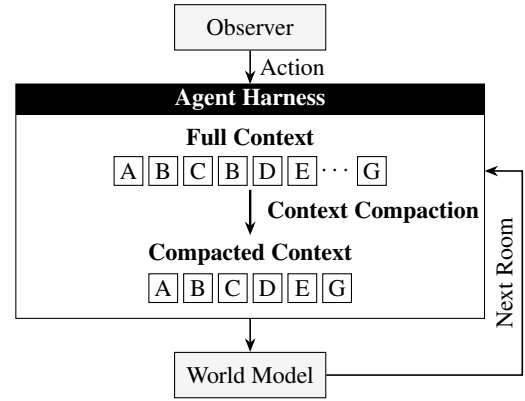
\begin{figure}[t]
  \centering
\begin{tikzpicture}[
    x=1cm, y=1cm, font=\footnotesize,
    box/.style={draw, fill=black!4, minimum height=0.6cm, minimum width=2.0cm, align=center},
    room/.style={draw, minimum width=0.38cm, minimum height=0.38cm, inner sep=0, fill=white},
    target/.style={room, fill=perlayer, text=white},
    flow/.style={-{Stealth[length=1.8mm, width=1.4mm]}, line width=0.6pt},
  ]
  \node[box] (observer) at (3.4, 0) {Observer};

  \draw (0.3, -0.75) rectangle (6.5, -3.85);
  \fill[black] (0.3, -0.75) rectangle (6.5, -1.15);
  \node[text=white] at (3.4, -0.95) {\textbf{Agent Harness}};

  \node at (3.4, -1.45) {\textbf{Full Context}};
  \foreach \i/\r in {0/A, 1/B, 2/C, 3/B, 4/D, 5/E} {
    \ifx\r A
      \node[target] at (3.4 + \i * 0.46 - 3.5 * 0.46, -1.9) {\r};
    \else
      \node[room] at (3.4 + \i * 0.46 - 3.5 * 0.46, -1.9) {\r};
    \fi
  }
  \node at (3.4 + 6 * 0.46 - 3.5 * 0.46, -1.9) {$\cdots$};
  \node[room] at (3.4 + 7 * 0.46 - 3.5 * 0.46, -1.9) {G};

  \draw[flow, line width=0.9pt] (3.4, -2.15) -- (3.4, -2.75);
  \node[anchor=west] at (3.5, -2.45) {\textbf{Context Compaction}};

  \node at (3.4, -3.0) {\textbf{Compacted Context}};
  \foreach \i/\r in {0/A, 1/B, 2/C, 3/D, 4/E, 5/G} {
    \ifx\r A
      \node[target] at (3.4 + \i * 0.46 - 2.5 * 0.46, -3.45) {\r};
    \else
      \node[room] at (3.4 + \i * 0.46 - 2.5 * 0.46, -3.45) {\r};
    \fi
  }

  \node[box] (model) at (3.4, -4.6) {World Model};
  \draw[flow] (observer.south) -- node[right] {Action} (3.4, -0.75);
  \draw[flow] (3.4, -3.85) -- (model.north);
  \draw[flow] (model.east) -- (7.0, -4.6) -- node[above, sloped] {Next Room}
    (7.0, -1.9) -- (6.5, -1.9);
\end{tikzpicture}
  \caption{Our agent harness conducts context compaction before sending it to a world model. It records the entire session as the full context, which soon
  outgrows the model's context window. Context compaction allows it to send the  model a walk through every visited room instead of the full context.}
  \label{fig:harness}
\end{figure}

Popular LLM agents such as Claude Code and Codex 
are responsible for managing the context of their underlying LLM~\citep{tirmazi2026compaction}. In the same way, we build an agent 
harness for our world models to manage context. Note that similar to LLM agent harnesses, our harness does not require the underlying world model to be retrained. Also like LLM agent harnesses, our harness is independent of the underlying model and can be used interchangeably with all four model designs that we test on and, in fact, should be easy to adapt for any world model that can work with our benchmark. Analogous to how an LLM along with the harness is called an LLM agent, we refer to a world model together with its harness as a \textit{world agent}.

Our agent harness is situated at the interface between a world model and 
an observer. We record every frame and action of an 
observer's session with the model in the harness. This is 
referred to as the \textit{full context} of the agent session. However, 
recall that the world model only has a context window length of 128 steps. 
As we mentioned earlier, LLM 
agent harnesses send a subset or summary of their full context to the underlying 
LLM using a process called context compaction. In the case of our world 
agent, without context compaction the original visit of any room revisit that has a gap larger 
than 128 cannot be retrieved from the context. However, our agent harness can send a \textit{compacted context} to the world model (Figure~\ref{fig:harness}).

\subsection{Compacted World}

Recall that an observer can choose one of four available actions in a given room. The observer can take a door at any of the room's four borders to go into a neighboring room. In the agent harness, we can build the compacted context of the session using only the observers actions. After an action, the observer moves 
by one room in our room grid. Therefore, we track the position of 
the observer in the harness without storing any information about the room's tiles. If the observer takes two steps at the same position, 
we only need to keep the most recent frame of the visited room in the context. 
Note that the harness cannot simply send these frames as a
list since our models trained only on raw steps, in which every step moves to an
adjacent room. The compacted context therefore has to look like a real walk. 

Our agent harness plans the shortest walk that passes through every visited room and ends at the current one. This is an instance of the traveling salesman problem. Since the traveling salesman problem is NP-hard in
general, our harness uses the Christofides approximation~\citep{christofides2022tsp} to
plan a short tour through every visited room and removes the longest edge of the tour to
obtain a walk. Each step of the walk carries the
newest frame of its room and the action toward the next room. For example, suppose the observer's trajectory in a $4$ room world is $A, B, A, B, C, B, C, D, C, D$. The compacted context is the walk $A, B, C,
D$. When the visited rooms do not lie on a single path, the walk must pass through
some rooms more than once. If room $B$ connects to rooms $A$, $C$, and $D$, for
example, the walk is $A, B, C, B, D$.

\subsection{Compaction Policies}

Our agent harness can follow one of two policies.
Under \textit{compact every step}, the harness compacts the full context before
every call to the world model. Under \textit{compact when full}, which is similar to the policy
of coding agents such as Claude Code\footnote{We say the policy is similar, not identical, because LLM agents often compact when the context is nearly full, e.g., at 95\% of the context window of the underlying LLM~\citep{tirmazi2026compaction}.}, the harness appends raw steps
until the context holds 128 steps, compacts it, and then appends raw steps again. During our experiments with the compact when full policy, we observed that the agent harness compacted about once every 50 steps. We also observed that the context sent to the world model held about 103 steps on average.

\section{Evaluation}

We begin by describing the experimental setting in \S~\ref{sec:experimental-setting}.
We then answer three questions that follow our contributions. First, how far
can the agent harness extend the context retrieval of a world model without retraining
it (\S~\ref{sec:eval_harness})? Second, how does the KV cache
design at the model layer trade memory against context retrieval
(\S~\ref{sec:eval_model})? Third, how does dreaming affect the context retrieval of each KV
cache design (\S~\ref{sec:dreaming})? Finally, we measure the agent latency of our world agents, split into the inference
latency of the world model and the latency of the agent harness
(\S~\ref{sec:eval_latency}). The source code for our work and our benchmark generator are available in our
\href{https://github.com/jadidbourbaki/context-compaction-world-models}{GitHub repository}.
The weights of all 21 trained world models are available on
\href{https://huggingface.co/jadidbourbaki/context-compaction-world-models}{Hugging Face}.

\subsection{Experimental Setting}\label{sec:experimental-setting}

We train our world models on $5000$ generated episodes of our benchmark world. 
Each episode is $512$ steps. We evaluate our models on $400$ separate episodes. Each training window spans $128$ steps. We train with $3$ random seeds, $42$, $43$, and $44$, and conduct 20K optimizer updates using batch size $4$. We give each world model identical training windows in the same order for each seed. We use AdamW~\citep{loshchilov2019adamw} for optimization with a peak learning rate of $3 \times 10^{-4}$, $\beta_1=0.9$, $\beta_2=0.95$, a weight decay of $0.1$, gradient norm clipping at 1.0, and a one-cycle schedule. During each evaluation, we sample 40 revisits from each of our $7$ gap bins. We score every world model and every context that the agent harness sends on the  same revisits. In our figures and results, we report the mean over our $3$ seeds and error bars are one population standard deviation.

\begin{figure}[t]
  \centering
  \includegraphics[width=\linewidth]{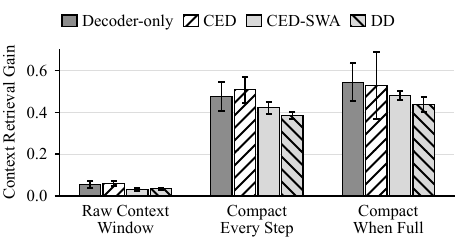}
  \caption{Performing context compaction at the agent harness layer lets every world model retrieve rooms beyond its raw
  context window. Each bar shows the context retrieval gain for revisits with gaps in the range $[128, 511]$ steps.}
  \label{fig:interaction}
\end{figure}

\subsection{Agent Harness Layer}
\label{sec:eval_harness}

In this section, we show that context compaction at the agent harness layer 
allows us to retrieve rooms whose original visit was up to 511 steps earlier. This is 4 times beyond the $128$ step context window size 
of our world models. This does not require 
any retraining. Figure~\ref{fig:interaction} shows 
the context retrieval gain for all four model designs for revisits 
with gaps in the range $[128, 511]$. As expected, since the context window length is only 128 steps, sending the raw context gives us 
a gain of near zero, specifically in the range $[0.03, 0.06]$. 
If we conduct context compaction at every step, the gain 
becomes significantly larger for all four models, lying in the range 
$[0.39, 0.51]$. If we only conduct context compaction when the context 
is full, the gain is in the range $[0.44, 0.54]$ for all models. We believe the compact when full policy scores slightly higher for two reasons: 1) its context holds more steps and keeps the most recent steps raw similar to the world model's training data, and 2) it is less affected by the five-step bias of the shuffled run (Appendix~\ref{app:gains}).  Compact when full also has a lower agent latency per step (\S~\ref{sec:eval_latency}). To rule out that these gains come simply from sending a shorter context, we send a truncated context that holds only the most recent steps, as many as the compacted context holds. This results in a near zero gain similar to the raw context window.

\begin{figure}[t]
  \centering
  \includegraphics[width=\linewidth]{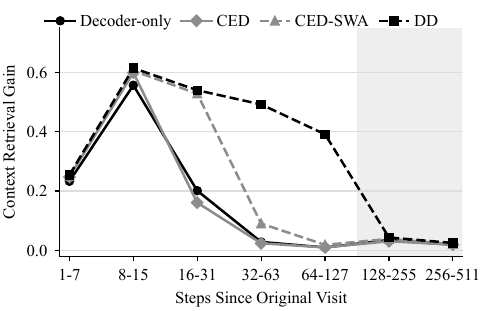}
  \caption{Context retrieval gain of our world model designs when the memory budget is fixed at 2,304 vectors. Note that each vector has $D = 384$ elements and we use $\rho = 32$ bits per floating point.}
  \label{fig:same_budget}
\end{figure}

\begin{figure}[t]
  \centering
  \includegraphics[width=\linewidth]{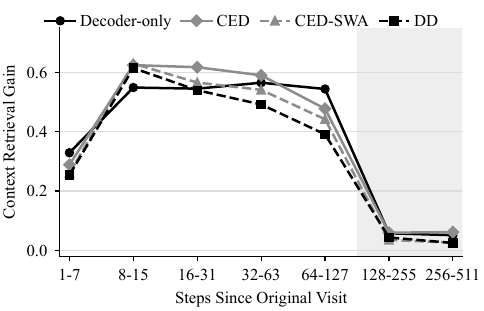}
  \caption{Context retrieval gain of our world model designs when the raw context window size is fixed at $128$ steps.}
  \label{fig:same_context}
\end{figure}

\subsection{Model Layer}
\label{sec:eval_model}

In this section, we show that when designing the KV cache at the model layer, world model
designers face a tradeoff between how much context fits in a given amount of memory and
how accurately the world model retrieves information from that context. We show both
sides of this tradeoff in Figures~\ref{fig:same_budget} and~\ref{fig:same_context}. In Figure~\ref{fig:same_budget}, we fix the memory budget of the world model's KV cache to 2,304 vectors. This memory budget is enough for us to store all 128 steps in the context window if we are using DD, since we only need to store 2 vectors per token for DD. However, since we need to store 16 vectors per token for Decoder-only and CED, we cannot store more than 16 steps in their context window. Similarly, since we need to store 8 vectors per token for CED-SWA, we cannot store more than 32 steps in its context window under our memory budget. Please refer to Appendix~\ref{app:budget} for more details. 

For our constrained memory budget experiment, we train bounded
versions of Decoder-only and CED in which their transformer blocks attend only to the 16 most recent steps. Similarly, we train a bounded version of CED-SWA, where 
the upper blocks attend only to the 32 most recent steps. We
train these bounded versions of our world models from scratch, with identical training data, seeds, and number of
updates as the existing (unbounded) versions of our world models. An alternative here would have been to shorten the context windows of existing versions of our world models only at
inference time to fit the memory budget. However, this approach would make the existing versions of the world models read shorter context windows compared to the ones they were trained on. This confounds the effect of smaller memory with the effect of the mismatch between the context lengths the world model trained on and the ones it is evaluated on making it difficult to isolate these two in our results. Note that we do not need to train a bounded version of DD because 
the unbounded 128-step context window version of DD already fits within our memory budget. Therefore, we use the same version of our DD world model in both Figure~\ref{fig:same_budget} and Figure~\ref{fig:same_context}.

Using the same memory budget, DD is able to retrieve rooms with revisit gaps of up to 127 steps. However, the context retrieval gain of Decoder-only, CED, and CED-SWA declines significantly as soon as the revisit gap becomes larger than their context window. In other words, when memory is the
constraint, a smaller KV cache per token lets a world model retrieve rooms much further back in a world agent exploration session.

In Figure~\ref{fig:same_context}, we instead fix the context length, i.e., all
four world models read the same 128-step raw context window. Here, the smaller KV
cache of DD comes at a price. DD retrieves rooms 8 to 31 steps back as well as Decoder-only, but
its context retrieval gain falls behind as the revisit gap grows, e.g., 0.39 compared
to 0.54 for revisit gaps in the range $[64, 127]$. Compared to Decoder-only, DD makes three changes to the architecture of the world
model: it computes the keys and values of its upper blocks from $z$, it uses SWA in its
lower blocks, and it shares a single KV cache across all of its upper blocks. To
understand which of these changes lowers the context retrieval gain, we compare
neighboring designs in Figure~\ref{fig:architecture}, since each design changes exactly
one property of the previous one. For revisit gaps of 32 to 63 steps, computing the upper keys and values from $z$, as in CED, does not lower
the context retrieval gain. Switching the lower blocks to SWA, as in CED-SWA, lowers
it. Sharing a single KV cache across all upper blocks, as in DD, lowers it
further.\footnote{For revisit gaps in the range $[32, 63]$, CED matches or exceeds
Decoder-only. DD falls below CED-SWA. CED-SWA falls below CED
in $2$ of $3$ seeds. In the third seed, Decoder-only and CED were still learning to
retrieve distant rooms when training stopped. Hence, we compare the designs within each seed.} In other words, the two changes that shrink the
KV cache of DD, i.e., its SWA lower blocks and its shared KV cache, are the ones that lower
context retrieval, while computing the keys and values from $z$ only saves prefill
computation (\S~\ref{sec:eval_latency}).

\subsection{Dreaming}
\label{sec:dreaming}

\begin{figure}[t]
  \centering
  \includegraphics[width=\linewidth]{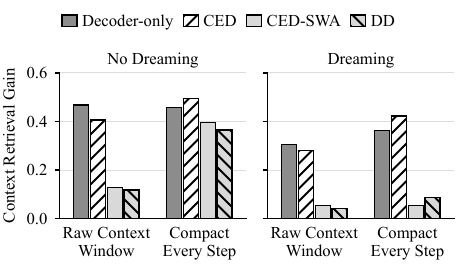}
  \caption{The effect of dreaming on context retrieval gain for our four world model designs.}
  \label{fig:revisits}
\end{figure}

The process of \textit{dreaming} occurs when we make a world model predict new frames based on frames previously
predicted by the world model~\citep{ha2018worldmodels,hafner2020dreamer}. Without
dreaming, we include the ground truth frames from our benchmark world in the world model's context window and only use the world model to predict the current frame. In other words, dreaming implies we are allowing the context window of a world model to include frames from the
world model's own output. Dreaming is often a more realistic setting for world agents
in the same way that LLM agents typically include their own outputs from previous prompts
in the session in their context.

In our regular evaluation, we store the true frame in the agent harness's session history and only use the world model to predict the current room. In the dreaming setting, we instead start storing the world model's predicted frames in the agent harness's
session history. Note that in our world agents, we store the layout of the benchmark world's room grid in the agent harness without actually storing the actual room, i.e., what tiles each room has. On an observer's first visit to a room, we store the true frame of the room in the agent harness's session history. Note that
we do not use the world model on a first visit. On a revisit, we use the world model to predict the room from the context we send it through the agent harness. We then score the world model's prediction
against the true room. In other words, our world agents only dream on revisits. In this section, we show that
world models whose lower blocks use SWA, i.e., CED-SWA and DD, lose almost all of their
context retrieval once they dream, whether the agent harness sends the raw context window
or a compacted context.

We run our world agents for sessions of 512 steps in newly generated
evaluation worlds that our world models were not trained on. In each session, the observer
alternates between a loop of 128 steps (i.e. a revisit gap of exactly 128 steps) and a random walk of 128 steps. We measure the context
retrieval gain on the revisit both with and without dreaming. Figure~\ref{fig:revisits} shows our results. Without dreaming, compact every step has
a gain in the range $[0.37, 0.50]$ for all four world models. With dreaming,
Decoder-only and CED keep a gain in the range $[0.28, 0.42]$ with either context.
However, the gain of CED-SWA and DD drops to at most 0.09. The drop also appears at short gaps. With loops of only 8 steps, the
gain of CED-SWA and DD with dreaming is at most 0.14, compared to at least 0.33 for
Decoder-only and CED (Appendix~\ref{app:closed_loop}).

Since CED and CED-SWA differ only in their lower blocks, which use SWA in CED-SWA, the
drop comes with the switch to SWA lower blocks. Note that our
experiments show where the drop appears, but they do not establish why SWA lower blocks
cause it. Consequently, the longer context that SWA buys DD at a fixed memory budget
(\S~\ref{sec:eval_model}) only helps a world agent that does not dream.

\subsection{Agent Latency}
\label{sec:eval_latency}

\begin{figure}[t]
  \centering
  \includegraphics[width=\linewidth]{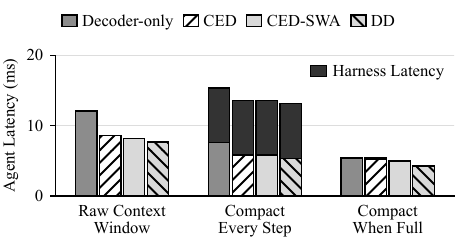}
  \caption{Mean agent latency per step over a 512-step exploration session of the benchmark
  world.}
  \label{fig:latency}
\end{figure}

Agent latency comprises the inference latency of the model and the agent harness's latency which is mostly context compaction time.  In Figure~\ref{fig:latency}, we show the mean agent latency per step of a 512-step session for our world models trained using seed 42. For the raw context window, the harness latency is negligible (about 8 $\mu$s per step). Also for the raw context window, we have to prefill 
the whole raw context window on every step after step 128. This is because the 
learned position embeddings we use in our world models are based on the position 
of a step in the context window, i.e., the first step of the window gets the first position embedding, the second
step gets the second one etc. Hence, when the context window is full and a new step
pushes the oldest step out of the window, all remaining steps are shifted by one position. Consequently, the keys and values in the KV cache, which were computed at the old
positions, no longer match their current steps. This forces the agent harness to prefill the entire window
again. In the compact every step policy, we plan a new walk before every step.
Since each new walk can visit the rooms in a different order, the context changes from one
step to the next forcing us to prefill the compacted context on every step of the
session. The
inference latency is lower than with the raw context window since the compacted context
itself is smaller. In the compact when full policy, we only plan a walk when the context is full, which
happens about once every 50 steps. Otherwise, we just decode a step from the KV
cache. Hence we only prefill on 8 of the session's 512 steps. Consequently, compact when full has the lowest agent latency. CED, CED-SWA, and DD prefill about 1.5 times faster than Decoder-only, since
their upper blocks only process the most recent step. Please see Appendix~\ref{app:timing} for further details on our agent latency evaluation.

\section{Related Work}

\citet{laird2025memory} compare memory mechanisms in transformer world models on a
state recall task. Our paper additionally matches parameter counts within 7\%
for our evaluated world models and measures context retrieval gain using a shuffled
run. R2M-Bench~\citep{gu2026r2mbench} compares each revisit with pairs from the same rollout
that revisit nothing at a matched gap. Its controls, however, still act on the output
of the world model. Our benchmark acts on the context instead. It rules out
context reconstruction by generating a new world for every episode and separates
context retrieval from context inference with the shuffled run.

At the model layer, we compare designs that share keys and values across blocks. Other work reduces the memory of the KV cache in different ways. StreamingLLM~\citep{xiao2024streamingllm} keeps
only a few initial tokens and a window of recent tokens. H2O~\citep{zhang2023h2o}
evicts all but the most important and the most recent tokens. KIVI~\citep{liu2024kivi} quantizes keys and values to 2 bits. Since these techniques act on
the KV cache of a given design, they can in principle be combined with the designs
we compare.

At the agent harness layer, MemGPT~\citep{packer2023memgpt} manages the context of an LLM by moving data
between its context window and an external memory, similar to how an operating system
manages memory. \citet{tirmazi2026compaction} formalize context compaction as the step in an agent loop
that shrinks the state of the agent to fit the context window of the model. In language, what a
compacted context preserves is hard to measure. In our benchmark world, however, we
know that a compacted context preserves every visited room, which lets us measure context retrieval through a compacted context
directly.

\section{Limitations and Conclusion}

Our work has three main limitations. First, we evaluate a single benchmark world at a
single model scale with three training seeds. Our world models also never train on
compacted contexts. Second, our world agents only dream on revisits and only for loops of
up to 128 steps. Therefore, we do not yet know how dreaming affects context retrieval
beyond the raw context window. Third, the rooms of our benchmark world are laid out on a predetermined grid which simplifies our context compaction algorithm. Consequently, our results show context compaction in its ideal case.
In world models where the camera moves continuously, an agent harness would first have to
decide which frames show the same place.

Context retrieval in world models depends on both layers of the inference
stack. At the model layer, a smaller KV cache per token reduces context retrieval
at a fixed context length but allows for a longer context at a fixed memory budget. At
the agent harness layer, context compaction lets a world model retrieve rooms far
beyond its context window without retraining. However,
once a world agent dreams, only the world models whose lower blocks attend over the full
context instead of relying only on SWA show consistent context retrieval, with or without context compaction. Therefore, it is often easier for a world
agent to extend the reach of its world model with context compaction at the agent
harness layer as opposed to sacrificing full attention in the lower blocks of the world model.

\bibliographystyle{mlsys2025}
\bibliography{references}

\section*{AI Usage Disclosure}

Generative AI was used as an assistant throughout this work for the following three
purposes. The first purpose was to help the authors write the code for 1) the benchmark, 2) the
world models, 3) the agent harness, and 4) the evaluation. Two agent harnesses, Claude Code and Codex, were also used to run the experiments under the
direction of the authors. The authors reviewed and corrected this code. The second purpose
was to help the authors draft and revise parts of the text, the figures, and the \LaTeX{}
code. The authors reviewed, rewrote, modified, and verified all of the text and take full responsibility for it. The third purpose was to
help the authors search for related work, which the authors then read and verified. Note that all of the content of this paper has been human reviewed and
verified for correctness prior to submission. The following LLMs were used for the above
purposes: Anthropic's Claude Opus 5.5, Claude Opus 5, and Claude Opus 4.8, and OpenAI's GPT-5.6 Sol and GPT-6.1 Sol. Finally, the authors also used the \textit{AI Mode} of Google's Search Engine to help with the initial research/exploration and web browsing that eventually led to the formulation of the paper's ideas.

\clearpage

\appendix

\section{Rigorous Background}
\label{app:background}

The decoder-only Transformer is one of the most popular designs for building
both world models and large language models. Let $\MD = U \circ \Norm \circ \Tau_N
\circ \cdots \circ \Tau_1 \circ E$ be a model with $N$ Transformer blocks
$\Tau_1, \ldots, \Tau_N$, an embedding layer $E$, a final normalization layer
$\Norm$, and an unembedding layer $U$.\footnote{Prior work also uses the term
layer for a transformer block. We use the two terms interchangeably.} Assume $\MD$
has a vocabulary size of $V$ with embedding width $D$. Consider a sequence of
$L$ input tokens. The embedding layer $E$ maps each input token to an embedding
vector of dimension $D$. Therefore, the shape of the input data becomes $[L,
D]$. Auxiliary data encoding the position or other information about the token
is also added to the embedding vector representing the token. Let $\tau_{0}$ be
the input data with auxiliary information added to it. For $i \geq 1$, let
$\iota_i$ and $\tau_i$ be the input and output of Transformer block $\Tau_i$
respectively. The model sends $\tau_{i - 1}$ as input to $\Tau_{i}$, i.e.,
$\iota_{i} = \tau_{i - 1}$ for every $i \geq 1$. In particular, $\Tau_1$
receives $\tau_{0}$. Note that all $\iota_i$ and $\tau_i$ tensors have the shape
$[L, D]$.

Consider any given Transformer block $\Tau_i$ in $\MD$. $\Tau_i$ processes its
input data first using self-attention and then using a multi-layer perceptron.
More concretely, $\Tau_i$ outputs $\tau_i \gets y + \MLP(\Norm[2](y))$ with $y
\gets \iota_i + \SA(\Norm[1](\iota_i))$ where $\MLP$ is a multi-layer
perceptron, $\SA$ is self-attention, and $\Norm[1]$ and $\Norm[2]$ are
normalization functions such as $\texttt{RMSNorm}$~\citep{zhang2019rmsnorm}. Note that the attention,
perceptron, and normalization parameters of each Transformer block $\Tau_i$ are
separate. $\MLP$ is a function $\R^{D} \to \R^{D}$, i.e., it is applied to each
of the rows of its input $\Norm[2](y)$ independently. Recall that in
$\tau_{0}$, the rows were the embedding vectors corresponding to the $L$ tokens.
This means that no information \textit{between} tokens is shared by $\MLP$.
$\MLP$ has a hidden width $D_{\mathrm{ff}}$ which is typically $4D$.
Consequently, in an $\MLP$, each of the $L$ rows is multiplied by a $[D,
D_{\mathrm{ff}}]$ matrix, passed through a non-linear function, and then 
multiplied by a
$[D_{\mathrm{ff}}, D]$ matrix.
The time complexity of one forward pass of $\MLP$ over the whole input is therefore $\bigO{L D
D_{\mathrm{ff}}} = \bigO{LD^2}$ (assuming $D_\mathrm{ff} = 4D$). Since $D$ is
fixed once the model is built, we can treat $D$ as a constant, giving us
$\bigO{L}$. $\SA$, on the other hand, is a function $\R^{L \times D} \to \R^{L
\times D}$ and information between tokens is shared inside $\SA$. In fact,
self-attention is the only part of the model $\MD$ that shares information between
tokens. We discuss self-attention in detail below.

One of the most common designs for self-attention is causal multi-head
self-attention. Assume our model $\MD$ uses self-attention with $h$ heads. Note
that $h$ needs to be a divisor of $D$. Let $h \cdot d_h = D$ for some $d_h$.
Each head is a function $\Head^{j} : \R^{L \times D} \to \R^{L \times d_h }$ for
$j \in \{1, \ldots, h\}$. Let $x$ denote the input to self-attention, a tensor
of shape $[L, D]$. Inside the Transformer block $\Tau_i$ this input is
$\Norm[1](\iota_i)$. Inside each head, $x$ is multiplied by $3$ learned
matrices $W_Q^{j}$, $W_K^{j}$, and $W_V^{j}$ each of which is of shape $[D,
d_h]$. They are called the query projection, the key projection, and the value
projection respectively. We then get $Q^{j} = x W_Q^{j}$, $K^{j} = x W_K^{j}$,
and $V^{j} = x W_V^{j}$ for each head $\Head^{j}$. Note that $Q^{j}$, $K^j$, and
$V^j$ are all matrices of shape $[L, d_h]$.

The final output of $\Head^{j}$ is then
\begin{equation}
  \Head^{j}(x) =
  \texttt{softmax}\left(\frac{Q^{j} (K^{j})\tran}{\sqrt{d_h}} + C\right) V^{j}.
  \label{eq:attention_head}
\end{equation}
Over here, $C$ is a matrix of shape $[L, L]$ such that $C_{st} = 0$ when $t \leq
s$ and $C_{st} = -\infty$ otherwise. The reason for adding $C$ is to ensure that
each token only gets information about itself and the tokens before it. This is
why $C$ is called the causal mask. Note that the softmax is taken along each row
of the matrix. The output of the softmax is therefore a matrix of shape $[L,
L]$. Since we then multiply it by $V^j$ which is a matrix of shape $[L, d_h]$,
the overall output of $\Head^{j}$ is a matrix of shape $[L, d_h]$. After
processing each head, we have a set of $h$ outputs $\Head^{j}(x)$. $\SA$
concatenates them to form
\begin{equation}
  \SA(x) = [ \Head^{1}(x) \parallel \cdots \parallel \Head^{h}(x) ] W_O.
  \label{eq:self_attention}
\end{equation}
Here $W_O$ is a learned matrix of shape $[D, D]$ called the
output projection. Note that the concatenated matrix has shape $[L, h \cdot d_h]
= [L, D]$, so $\SA(x)$ is a matrix of shape $[L, D]$. This is what makes the
residual addition $\iota_i + \SA(\Norm[1](\iota_i))$ well defined, and why
$\tau_i$ is a matrix of shape $[L, D]$ as discussed earlier. The mask $C$
defined above gives each token access to its entire prefix. Some architectures
use sliding window attention, as opposed to ordinary (i.e., full) attention.
Sliding window attention restricts the self-attention of a Transformer block to
a bounded window of $w$ tokens. It replaces $C$ in
Equation~\ref{eq:attention_head}
with the windowed causal mask $C^{w}$, where $C^{w}_{st} = 0$ when $s - w < t
\leq s$ and $C^{w}_{st} = -\infty$ otherwise. Other than the mask, a Transformer
block that uses sliding window attention is identical to a regular Transformer
block.

Consider a given attention head $\Head^{j}$ in $\SA$ of a Transformer block $\Tau_i$. The
query, key, and value projections in $\Head^{j}$ are matrix multiplications of an
$[L, D]$ matrix and a $[D, d_h]$ matrix. The time complexity of the three
projections in $\Head^{j}$ is therefore $\bigO{L D d_h}$. Since $\Tau_i$ has $h$ heads,
the time complexity of each projection over all heads is $\bigO{h L D d_h} = \bigO{L D^2}$ 
because $h \cdot d_h = D$. Note that this only accounts for the projections 
and not the subsequent attention computations. For the attention computation, 
$Q^{j} (K^{j})\tran$ is a multiplication between an $[L, d_h]$ matrix and a 
$[d_h, L]$ matrix so it costs $\bigO{L^2 d_h}$ per head, i.e.,
$\bigO{L^2 D}$ over all heads. Multiplying the softmax output by $V^j$ is
a multiplication between an $[L, L]$ matrix and an $[L, d_h]$ matrix so it
costs $\bigO{L^2 d_h}$ per head, i.e., $\bigO{L^2 D}$ over all heads.
Finally, the output projection $W_O$ is a multiplication between 
an $[L, D]$ matrix and a $[D, D]$ matrix so it costs $\bigO{L D^2}$. 
The overall time complexity of one forward pass of $\SA$ is therefore 
$\bigO{L D^2 + L^2 D}$. Similar to $\MLP$, we can assume $D$ is constant, 
which simplifies the time complexity to $\bigO{L^2}$. Recall that 
$\Tau_i$ computes $\SA$ followed by $\MLP$. The overall time 
complexity of one block $\Tau_i$ is therefore $\bigO{L^2}$. Since 
the model $\MD$ has $N$ Transformer blocks, the overall time complexity of one 
forward pass of $\MD$ is $\bigO{N L^2}$.

\subsection{Inference and KV Cache}
\label{sec:kv_cache}

During inference, a decoder-only Transformer $\MD$ generates
output tokens sequentially. Consider a model $\MD$ with the parameters
as defined earlier. Let's assume $\MD$, upon receiving a sequence
of $L$ input tokens,
produces output tokens one at a time. To process the attention
head $\Head^{j}$ in $\Tau_i$, we need to compute the queries, keys, and values for
the relevant tokens. Specifically, we first focus on the computation
of any given row $r$ in $\Head^{j}(x)$. To get row $r$ of $\Head^{j}(x)$,
we need $q_r$ together with $k_1, \ldots, k_r$ and $v_1, \ldots, v_r$. Here,
$q_r$, $k_r$, and $v_r$ are row $r$ of the
query projection $Q^{j}$, key projection $K^{j}$,
and value projection $V^{j}$, respectively. This follows directly 
from the fact that we first have to multiply $Q^{j}$ and $(K^{j})\tran$ 
inside the softmax function of $\Head^{j}(x)$. The softmax function itself 
is computed row by row. After the softmax, we
multiply the softmax output by the value projection $V^{j}$ which means
we need $v_1, \ldots, v_r$ as well. Note that we do not need 
$k_{r+1}, \ldots, k_L$ and $v_{r+1}, \ldots, v_L$  because the 
causal mask $C$ ensures that the attention weights for these positions are zero.

Since $W_Q^j$, $W_K^j$, and $W_V^j$ are $[D, d_h]$ matrices, and
a row of $x$ is a $[1, D]$ vector, computing their products
$q_r$, $k_r$, and $v_r$ for some row $r$ is $\bigO{D d_h}$.
Recall that we assume $D$ and $d_h$ are constant, making the computation of
$q_r$, $k_r$, and $v_r$ effectively $\bigO{1}$. Once $k_1, \ldots, k_r$ and
$v_1, \ldots, v_r$ are available, row $r$ of $\Head^{j}(x)$ performs $r$ dot
products and a weighted
sum of $r$ values. Each of them is $\bigO{1}$, so row $r$ costs $\bigO{r}$.

If we use no form of caching, model $\MD$ has to recompute every row of
$\Head^{j}(x)$ to process the next token. This means recomputing every row of
$\Head^{j}(x)$ in every attention head $\Head^{j}$ of every Transformer block $\Tau_i$.
Computing all $L$ rows of a given attention head 
has time complexity $\sum_{r = 1}^{L} \bigO{r} = \bigO{L^2}$. Since the number
of heads $h$ is fixed, computing all $L$ rows for 
the entire Transformer block $\Tau_i$ also has time complexity $\bigO{L^2}$. 
Since model $\MD$ has $N$ Transformer blocks, the overall time complexity
of producing the next token after $L$ tokens without a cache is $\bigO{N L^2}$.

KV caches optimize inference by storing the key and value projections
for all processed tokens
in a given token sequence for each attention head $\Head^{j}$ of each
Transformer block $\Tau_i$. This means that to compute row $r$ of $\Head^{j}(x)$,
we can read $k_1, \ldots, k_{r-1}$ and $v_1, \ldots, v_{r-1}$ from the KV
cache and only compute $q_r$, $k_r$, and $v_r$ for the current row $r$.
With a KV cache, to compute the next token after a sequence of $L$ tokens,
we only compute $q_{L+1}$, $k_{L+1}$, and $v_{L+1}$ in each block which is $\bigO{1}$.
The dot products and the weighted sum then read $L + 1$ cached vectors 
which is $\bigO{L}$. Therefore, the time complexity of processing 
a new token is $\bigO{L}$ in each Transformer block $\Tau_i$ and $\bigO{N L}$ in the
entire model $\MD$. The KV cache therefore reduces the cost of generating
each new token from
$\bigO{N L^2}$ to $\bigO{N L}$. If a Transformer block $\Tau_i$ uses
sliding window attention, the time complexity of generating a new token
using $\Tau_i$ can be further reduced from $\bigO{L}$ to $\bigO{w}$,
where $w$ is the window size.

However, by using a KV cache, we are effectively reducing the time of 
processing new tokens by trading it off with increased memory usage. 
For a given attention head $\Head^{j}$ and a sequence of $L$ 
tokens, a KV cache stores $k_1,
\ldots, k_L$ and $v_1, \ldots, v_L$, which are $2L$ vectors of $d_h$ numbers
each. A Transformer block $\Tau_i$ has $h$ heads and $h \cdot d_h = D$. Therefore,
a KV cache stores $2LD$ floating point values for each Transformer block $\Tau_i$.
Since a model $\MD$ has $N$ Transformer blocks, the total 
size of its KV cache is $2NLD\rho$ where $\rho$ is the number of 
bytes per floating point value. If a given Transformer block $\Tau_i$ uses sliding 
window attention, its KV cache only needs to store $2wD\rho$ bytes, 
where $w$ is the window size.

Inference on a new prompt is typically done in two phases called the prefill phase and the 
decode phase. In the prefill phase, the model $\MD$
reads the $L$ input tokens in one forward pass and fills the KV
cache of every block with the key and value projection corresponding to each input token.
The prefill phase therefore has time complexity $\bigO{N L^2}$ 
even when the KV cache is used because it starts with an empty KV cache. 
In the decode phase, $\MD$ produces one new
token at a time while reading the KV cache and adding the new token's key and
value projections back to the KV cache. Hence the decode phase 
has time complexity $\bigO{N L}$ per token.

\subsection{Decoder-Decoder Models}

Prior work~\citep{sun2024yoco} proposes a novel architecture for large language
models called a Decoder-Decoder model. Let $\MDD$ be a Decoder-Decoder model.
$\MDD$ uses
$N = N_{\mathrm{s}} + N_{\mathrm{c}}$ Transformer blocks where the first
$N_{\mathrm{s}}$ blocks
are the lower blocks of $\MDD$ and the remaining $N_{\mathrm{c}}$ blocks
are the upper blocks of
$\MDD$. The lower blocks use a self-attention whose KV cache stays constant in
$L$. The lower blocks of a Decoder-Decoder model can be built with either sliding
window attention or gated retention. In the sliding window attention version, each
lower block attends over a window of $w$ tokens and therefore stores $2wD$
numbers regardless of $L$. Let $z = \Norm[3](\tau_{N_{\mathrm{s}}})$ be the normalized
output of the last lower block, where $\Norm[3]$ is a normalization function.
$\MDD$ projects $z$ once into a shared key
matrix $K^{\star} = z W_K^{\star}$ and a shared value matrix $V^{\star} = z
W_V^{\star}$, using learned matrices $W_K^{\star}$ and $W_V^{\star}$ of shape
$[D, D]$, so $K^{\star}$ and $V^{\star}$ both have shape $[L, D]$. The
$N_{\mathrm{c}}$ upper blocks
$\Tau_i$ of $\MDD$ use cross-attention $\CA$ as their token mixer.

In cross-attention, $K^{\star}$ and $V^{\star}$ are divided into slices
of $d_h$ columns where $h \cdot d_h = D$ and $h$ is the number of attention
heads per Transformer block. Let $K^{\star j}$ and $V^{\star j}$ denote the $j$-th
slices of $d_h$ columns of $K^{\star}$ and $V^{\star}$. Head $j$ of $\CA$
in an upper block $\Tau_i$ computes its own query projection $Q^{j} = x W_Q^{j}$.
Then, it uses the shared key and value matrices to compute the cross-attention
\begin{equation}
  \CA^{j}(x) = \texttt{softmax}\left(
    \frac{Q^{j} (K^{\star j})\tran}{\sqrt{d_h}} + C \right) V^{\star j}.
  \label{eq:cross_attention}
\end{equation}
Note that the mask $C$ in Equation~\ref{eq:cross_attention} is the full causal
mask of shape $[L, L]$ defined earlier. Each upper block therefore reads the
entire prefix through the shared pair. Each lower block reads only its window of
$w$ tokens. $\MDD$ remains autoregressive. Like $\SA$,
$\CA$ concatenates its $h$ head outputs and multiplies the result by 
an output projection $W_O$ of shape $[D, D]$. 
The only token-mixing parameters an upper block owns are therefore
$W_Q^{1}, \ldots, W_Q^{h}$ and $W_O$. The upper Transformer blocks 
do not have their own key or value projections.
This simplifies the KV cache to a single shared pair which holds $2LD$ 
numbers for any number of upper Transformer blocks. The total size of a 
KV cache for the entire model $\MDD$ is therefore 
$2LD\rho + 2 N_{\mathrm{s}} w D \rho$ bytes 
where the second term is a constant in $L$ and $\rho$ is the number of bytes
per floating point.
Compared to the KV cache of $\MD$,
which needs $2NLD\rho$ bytes, a Decoder-Decoder model divides the part of the
cache that grows with the history by $N$.

\subsection{Causal Encoder-Decoder Models}

DeepSeek recently released a large language model called 
DeepSeek-V4.1-Flash~\citep{deepseek2026ced}. DeepSeek-V4.1-Flash 
uses a novel architecture called a Causal Encoder-Decoder which 
is a descendant of Decoder-Decoder models.
Let $\MCED$ be a Causal Encoder-Decoder
model. $\MCED$ splits its $N = N_{\mathrm{s}} + N_{\mathrm{c}}$
Transformer blocks similar to Decoder-Decoder models. $\MCED$'s 
$N_{\mathrm{s}}$ lower
blocks are called encoder blocks and $\MCED$'s $N_{\mathrm{c}}$ upper 
blocks are called decoder blocks, respectively.
$\MCED$'s lower blocks, i.e., the encoder blocks, are ordinary decoder-only
Transformer blocks akin to those in $\MD$.
Each encoder block in $\MCED$ computes separate key and value projections 
from its input and maintains a dedicated KV cache. The output
of the final lower block of $\MCED$ is $z =
\Norm[3](\tau_{N_{\mathrm{s}}})$ where $\Norm[3]$ is a normalization function.
$\MCED$'s upper Transformer blocks, however, behave 
differently from Decoder-Decoder models. Unlike Decoder-Decoder models, which 
use a shared key and value matrix, each upper Transformer block $\Tau_i$ 
in $\MCED$ has its own key projection $W_K^{(i)}$ and value projection
$W_V^{(i)}$. Note that both $W_K^{(i)}$ and $W_V^{(i)}$ are learned matrices of 
shape $[D, D]$. A given decoder Transformer block $\Tau_i$ in $\MCED$ 
computes 
$K^{(i)} = z W_K^{(i)}$ and $V^{(i)} = z W_V^{(i)}$ which
are both matrices of shape $[L, D]$. Note that all $N_{\mathrm{c}}$
pairs of $K^{(i)}$ and $V^{(i)}$ are computed independently from the same $z$.
The description above follows the general formulation in the DeepSeek-V4.1-Flash technical report.
DeepSeek-V4.1-Flash adds components that $\MCED$ omits. Its first two encoder
layers use sliding-window attention. Its remaining encoder layers use compressed
sparse attention. Its decoder layers also run sliding-window self-attention over
their own hidden states. One of its decoder layers computes a global pair that
the other decoder layers reuse.

In an $\MCED$, each decoder Transformer block $\Tau_i$ uses its key and value projections
$K^{(i)}$ and $V^{(i)}$ to compute cross-attention. Let $K^{(i) j}$ and $V^{(i)
j}$ denote the $j$-th slices of $d_h$ columns of $K^{(i)}$ and $V^{(i)}$. Head $j$
of $\CA$ in decoder block $\Tau_i$ first computes the query projection
$Q^{j} = x W_Q^{j}$. Then head $j$ computes cross-attention $\CA^{j}(x)$ as follows.
\begin{equation}
  \CA^{j}(x) = \texttt{softmax}\left(
    \frac{Q^{j} (K^{(i) j})\tran}{\sqrt{d_h}} + C \right) V^{(i) j},
  \label{eq:ced_attention}
\end{equation}
where $C$ is the causal mask and $W_O$ is the output projection. Identically to
self-attention, the $h$ head outputs are concatenated before being multiplied by
$W_O$.
Each decoder Transformer block $\Tau_i$ in $\MCED$ therefore has its
own $W_Q^{1}, \ldots, W_Q^{h}$,
$W_O$, $W_K^{(i)}$, and $W_V^{(i)}$. The KV cache for
each decoder block $\Tau_i$ therefore has to store $2LD$ floating point values. 
The encoder blocks of $\MCED$ are identical to the Transformer blocks in $\MD$
so the KV cache of each encoder block has the same size as in $\MD$.
Hence the total size of the KV cache of $\MCED$ is $2NLD\rho$ bytes
where $\rho$ is the number of bytes per floating point value. 
Recall that this is the same as the total KV cache size of $\MD$. A Causal
Encoder-Decoder model, therefore, does not provide any KV cache memory savings
out of the box. However, $\MCED$ significantly reduces the amount of 
computation needed for prefill.
Recall from \S~\ref{sec:kv_cache} that $\MD$ fills the KV cache for a prompt
of $L$ tokens with one forward pass over all $N$ blocks in $\bigO{N L^2}$. In
$\MCED$ every $K^{(i)}$ and $V^{(i)}$ is a function of $z$. Hence 
the KV cache for a prompt is filled by executing the $N_{\mathrm{s}}$ 
encoder blocks and the $N_{\mathrm{c}}$ projections of $z$. 
Importantly, the decoder blocks only need to run for the last token of the prompt
during the prefill phase because filling their KV caches 
only needs projections of $z$. The prefill phase in $\MCED$
therefore costs $\bigO{N_{\mathrm{s}} L^2 + N_{\mathrm{c}} L}$ as compared 
to $\bigO{N L^2}$ in $\MD$. The decode phase has the 
same time complexity, $\bigO{N L}$, in both $\MD$ and $\MCED$ since every 
Transformer block still
has to compute the new token's query projection and attend over the $L$ cached tokens.

\section{Further Evaluation}
\label{app:detailed-eval}

This section contains a set of fun deep dives across multiple aspects of our evaluation.

\subsection{Agent Harness Layer}
\label{app:gains}

\begin{figure*}[t]
  \centering
  \includegraphics[width=\textwidth]{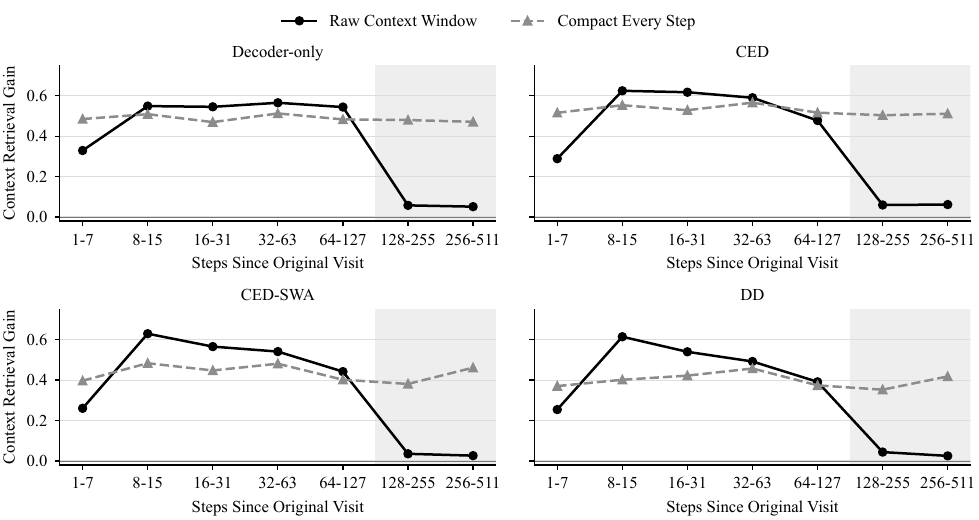}
  \caption{The context retrieval gain of all four of our world model designs
  over a range of gap bins. We use solid black lines to show the 128-step raw context window and
  use dashed gray lines to show a compacted context of about 80 steps.}
  \label{fig:compaction}
\end{figure*}

In Figure~\ref{fig:interaction} of \S~\ref{sec:eval_harness}, we showed 
the context retrieval gain for all four world model designs only over the 
gap bins that are larger than 128 steps, i.e., larger than the context window length 
of the models. In this section, we look at the context retrieval gain across all 
gap bins in detail. In Figure~\ref{fig:compaction}, we show the gain of each 
world model for all our gap bins both when we send the raw context window 
directly and when we conduct context compaction at each step. When we send 
the raw context window, the gain of all our world model designs drops to near zero 
as soon as the revisit gap is larger than the model's context window length of 128 steps. However, when we do context compaction, the gain stays 
roughly flat across all gap bins. Note that at gaps of 8 to 127 steps, the context compaction version almost always has a lower score compared to the raw context
window for all our world models. 
We suspect that part
of this difference comes from the shuffled run. The shuffled run keeps the five most
recent steps of the context, and a compacted walk can pass through the revisited room
within its last five steps. In that case, the shuffled context still contains the
revisited room, which lowers the context retrieval gain of the compacted context.

\subsection{Truncated Contexts}
\label{tab:truncated-contexts}

\begin{figure}[t]
  \centering
  \includegraphics[width=\linewidth]{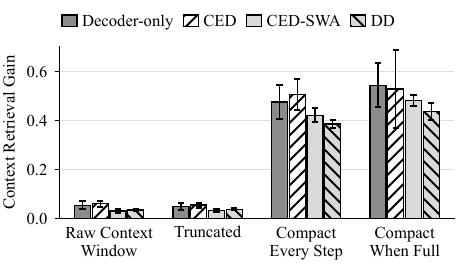}
  \caption{Context retrieval gain for gap bins of 128 to 511 steps.}
  \label{fig:contexts}
\end{figure}

As mentioned in \S~\ref{sec:eval_harness}, we want to rule out that the 
gains of context compaction come from simply sending a shorter context to the 
world model. 
In Figure~\ref{fig:contexts}, we compare context compaction with the alternative 
of simply truncating the context by keeping the most recent steps. Beyond 
128 steps, the context retrieval gain of using a truncated context drops 
to near zero similar to using no context compaction. Therefore, a shorter 
context alone is not the reason for context compaction improving a world 
model's long context retrieval beyond its context window length. 

\begin{figure*}[t]
  \centering
  \includegraphics[width=\textwidth]{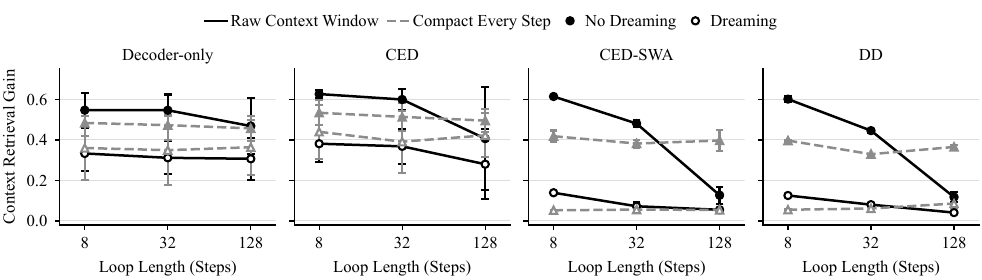}
  \caption{Context revisit gain for revisit loops of length $8$, $32$, and 
  $128$ steps respectively with and without letting the world model dream.}
  \label{fig:agent_sweep}
\end{figure*}

\subsection{Dreaming}
\label{app:closed_loop}

In Figure~\ref{fig:revisits} in \S~\ref{sec:dreaming}, we show the context 
retrieval gain in the dreaming setting for a loop with $128$ steps. In this 
section, we extend our results to loops of lengths 8 and 32 steps respectively.
Figure~\ref{fig:agent_sweep} shows our results. Without
dreaming and with the raw context window, the gain of CED-SWA and DD falls as the
loop grows since the starting room moves further back in the raw context window until it
is the oldest step at 128 steps. Without dreaming and with context compact at every step, the gain
of every world model stays roughly constant across loop lengths. With dreaming, the
gain of CED-SWA and DD stays at 0.14 or less at every loop length and with
either context. 

\subsection{Memory Budget}
\label{app:budget}

In Figure~\ref{fig:same_budget} in \S~\ref{sec:eval_model}, we use a fixed memory budget of $2,304$ vectors for the KV cache of all four of our world model designs. Note that each vector has 384 elements each of which is a 32-bit floating point value. In this section, we discuss the reasoning behind our memory budget. We first show that to maintain a context window size of 128 steps, our most memory efficient world model design, DD, requires $2,304$ vectors. As mentioned in 
\S~\ref{sec:eval_model}, we then show that the same memory budget results in a 
context window length of only 16 steps for Decoder-only and CED and a context window length of 32 steps for CED-SWA. Recall from
\S~\ref{sec:designs} that each step of our benchmark world is 9 tokens, i.e., 8 tokens for
the tiles of a room and 1 token for the action. In DD, we store a shared key vector and a
shared value vector per token for its upper blocks, i.e., 2 vectors per token. Therefore,
a context window of 128 steps needs $2 \times 9 \times 128 = 2{,}304$ vectors in DD. Since each
vector has $D = 384$ elements stored as 32-bit floating point values, this is about 3.5 MB. When using Decoder-only and CED, we have to store a key vector and a value vector per token in
each of their 8 transformer blocks, i.e., 16 vectors per token. Hence, their context
window is $2{,}304 / (16 \times 9) = 16$ steps. With CED-SWA, we have to store a key vector and a value
vector per token in each of its 4 upper blocks, i.e., 8 vectors per token. This results in a context
window size of $2{,}304 / (8 \times 9) = 32$ steps. Note that we also need to store  $4 \times 2 \times 9 = 72$ vectors for the $4$ lower blocks of CED-SWA
and DD that use SWA with 2-step windows. However, since the number of vectors needed by SWA blocks do not grow with the
context, we do not count them toward the budget for simplicity. They add only a fixed cost of about 3.125\% to the total memory of
CED-SWA and DD.

\subsection{Timing Methodology}
\label{app:timing}

\begin{table}[t]
  \centering
  \footnotesize
  \setlength{\tabcolsep}{3pt}
  \caption{The latency (in milliseconds) for a call to our world model designs using a context of 128 steps. We report the median latency over 30 iterations.}
  \label{tab:systems}
  \begin{tabular}{@{}lrrr@{}}
    \toprule
    & \multicolumn{2}{c}{Decode} & \\
    \cmidrule(lr){2-3}
    Model & Batch 1 & Batch 16 & Prefill \\
    \midrule
    Decoder-only & 5.74 & 17.42 & 14.53 \\
    CED       & 5.89 & 18.06 & 9.80 \\
    CED-SWA   & 5.30 & 11.37 & 9.56 \\
    DD        & 4.49 & 7.53  & 9.20 \\
    \bottomrule
  \end{tabular}
\end{table}

In this section, we look a bit more carefully into the inference latency 
of our world model designs. In Table~\ref{tab:systems}, we show 
the decode latency and prefill latency of all four of our 
world model designs. Decode latency, as is standard terminology, is the time to predict one step when we already hold the previous 127 steps in the KV cache. We measure decode latency at batch sizes of 1 and 16. Prefill latency is the
time to read the 128-step raw context window, fill them up in an empty KV cache, and then predict the next step. Note that we only need to run prefill on the
upper blocks of CED, CED-SWA, and DD at the last step (step 128) since their upper blocks only see
earlier steps through the projected keys and values. We benchmark the trained world models of seed 42, i.e., the same world models as in
Figure~\ref{fig:teaser}, on frames of our evaluation recording. We run all of our measurements in float32 on the GPU of an Apple M4 Pro with
PyTorch 2.14. In Table~\ref{tab:systems}, we show the median latency over 30 iterations of the same request to the world model. To
measure the agent latency per step in Figure~\ref{fig:latency}, we replay the first episode
of our evaluation recording under each policy. For each step in the episode, we record 1)
whether the policy decodes one step or prefills its context, 2) how many steps the context
holds, and 3) the harness latency, i.e., how long the agent harness takes to build the
context sent to the world model. We measure the harness latency of a step as the median
over 3 runs of building its context, i.e., planning and gathering a compacted context or
stacking the frames of the raw context window. We measure the inference latency of a step
as the median over 10 runs of the world model decoding or prefilling a context of that
length. The agent latency of a step is the sum of its harness latency and its inference
latency. Note that in the compact every step policy (the policy where context compaction occurs the most), the harness latency is 7.8 ms per step on average and
15.0 ms at most.

\end{document}